\documentclass[runningheads,a4paper]{llncs}

\usepackage{amssymb}
\usepackage{amsmath}
\usepackage{graphicx}

\usepackage{url}
\urldef{\mailsa}\path|{alfred.hofmann, ursula.barth, ingrid.haas, frank.holzwarth,|
\urldef{\mailsb}\path|anna.kramer, leonie.kunz, christine.reiss, nicole.sator,|
\urldef{\mailsc}\path|erika.siebert-cole, peter.strasser, lncs}@springer.com|    

\usepackage{color}

\begin{document}

\mainmatter  % start of an individual contribution

% first the title is needed
\title{Fast CapsNet for Lung Cancer Screening \vspace{-4mm}}
\author{Aryan Mobiny, Hien Van Nguyen \vspace{-2mm}}
\institute{Department of Electrical and Computer Engineering\\ University of Houston}
\maketitle

\begin{abstract}
\vspace{-6mm}
Lung cancer is the leading cause of cancer-related deaths in the past several years. A major challenge in lung cancer screening is the detection of lung nodules from computed tomography (CT) scans. State-of-the-art approaches in automated lung nodule classification use deep convolutional neural networks (CNNs). However, these networks require a large number of training samples to generalize well. This paper investigates the use of capsule networks (CapsNets) as an alternative to CNNs. We show that CapsNets significantly outperforms CNNs when the number of training samples is small. To increase the computational efficiency, our paper proposes a consistent dynamic routing mechanism that results in $3\times$ speedup of CapsNet. Finally, we show that the original image reconstruction method of CapNets performs poorly on lung nodule data. We propose an efficient alternative, called convolutional decoder, that yields lower reconstruction error and higher  classification accuracy.

\end{abstract}

\section{Introduction}
Lung cancer is consistently ranked as the leading cause of the cancer-related deaths all around the world in the past several years, accounting for more than one-quarter (26\%) of all cancer-related deaths \cite{siegel2017cancer}.  The stage at which diagnosis is made largely determines the overall prognosis of the patient. The five-year relative survival rate is over 50\% in early-stage disease, while survival rates drop to less than 5\% for late-stage disease \cite{siegel2017cancer}. A major challenge in lung cancer screening is the detection of lung nodules \cite{national2011reduced}.

Convolutional neural networks (CNNs) \cite{krizhevsky2012imagenet,he2016deep} are the state-of-the-art methods in lung nodule classification. However, CNNs have a number of important drawbacks mostly due to their procedure of routing data. Routing is the process of relaying the information from one layer to another layer. CNNs perform routing via pooling operations such as  max-pooling and average-pooling. These pooling operations discard information such as location and pose of the objects which can be valuable for classification purposes.

Recently, Sabour et al. \cite{sabour2017dynamic} introduced a new architecture, called Capsule Network, to address CNNs' shortcomings. The idea is to encode the relative relationships (e.g., locations, scales, orientations) between local parts and the whole object. Encoding these relationships equips the model with a built-in understanding of the 3D space. This enables CapsNet to recognize objects from different 3D views that were not seen in the training data. 

CapsNets employ a dynamic routing mechanism to determine where to send the information. Sabour et al. \cite{sabour2017dynamic} successfully used this algorithm for training the network on hand-written images of digits (MNIST) and achieved state-of-the-art performance. However, it remains unclear how CapsNets perform on medical image analysis tasks. In addition, the dynamic routing operations are computationally expensive making them significantly slower than other modern deep networks. This prevents the use of CapsNet for higher dimensional data such as 3D computed tomography (CT) scans. It is also unclear how CapsNet will compare to state-of-the-art methods on medical imaging tasks. 

Motivated by these observations, this paper makes the following contributions: i) We investigate the performance of CapsNet on 3D lung nodule classification task. We show that CapsNets compare favorably to CNNs when the training size is large, but significantly outperform CNNs for small size datasets. ii) We propose a consistent dynamic routing mechanism to speed up CapsNets. The proposed network runs  $3\times$ faster than the original CapsNets on 3D lung data. iii) We develop an efficient method, called convolutional decoder, that is more powerful in reconstructing the input from the final capsules. The proposed decoder serves as a regularizer to prevent over-fitting issue, and yields higher classification accuracy than the original CapsNets.

\section{Capsule Network}
\begin{figure}[t]
    \centering
    \includegraphics[width=\textwidth]{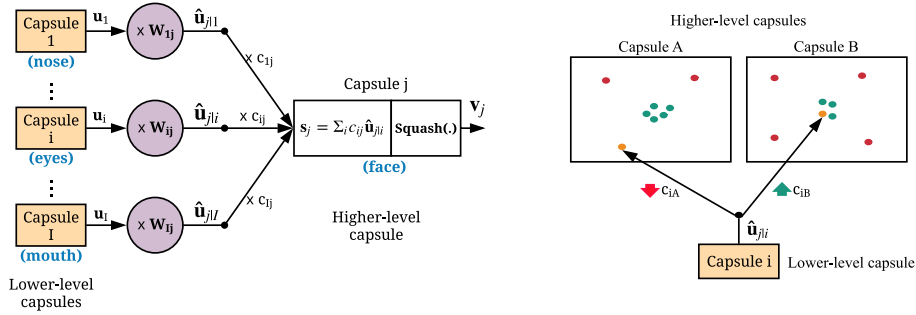}
    \caption{\textbf{Left:} connections between the lower and higher-level capsules , \textbf{Right:} dynamic routing for sending information from a lower-level capsule to higher-level ones.}
    \label{capsule}
\end{figure}

\textbf{Capsule Computation}: In this architecture, a capsule is defined as a group of neurons (whose outputs form an \textit{activation vector}). They predict the presence and the pose parameters of a particular object at a given pixel location. The direction of an activation vector captures the object's pose information, such as location and  orientation, while the length (a.k.a norm or magnitude) of the activation vector represents the estimated probability that an object of interest exists. For instance, if we rotate an image, the activation vectors also change accordingly, but their lengths stay the same. Fig. \ref{capsule} illustrates the way CapsNets route the information from one layer to another layer, using face detection as an example. Lengths of the outputs of the lower-level capsules ($\mathbf{u}_1,\mathbf{u}_2,\ldots, \mathbf{u}_I$) encode the existence probability of their corresponding entity (e.g. eyes, nose, and mouth). Directions of the vectors also encodes various properties of a particular entity, such as its size, orientation, position, etc.

The relationship between $i$-th capsule in a lower layer and $j$-th capsule in the next higher layer is encoded using a linear transformation matrix $\mathbf{W}_{ij}$. The information is propagated as:  $\hat{\mathbf{u}}_{j|i} = \mathbf{W}_{ij} \mathbf{u}_i$. The vector $\hat{\mathbf{u}}_{ij}$ represents the belief of $i$-th capsule in a lower layer about $j$-th capsule in the higher layer. In our example, $\mathbf{\hat{u}}_{j|1}$ represents the predicted pose of the face according to the detected pose of the nose. During the training, the network will gradually learn a transformation matrix for each capsule pair to encode the corresponding part-whole relationship. 

\noindent
\textbf{Dynamic Routing}: Having computed the prediction vectors, the lower-level capsules then route their information to parent capsules that agree the most with their predictions. The mechanism that ensures that the outputs of the child capsules get sent to the proper parent capsules is named \textit{Dynamic Routing}. Let $c_{ij}$ denotes the routing coefficient from $i$-th capsule in the lower layer to $j$-th capsule in the higher layer, where $\sum_j c_{ij} = 1$ and $c_{ij} \ge 0, \;\; \forall j$. When $c_{ij} = 1$, all information from $i$-capsule will be sent to $j$-capsule, whereas when $c_{ij}=0$, there is no information flowing between the two capsules. Dynamic routing method iteratively tunes the $c_{ij}$ coefficients and routes the child capsules' outputs to the appropriate capsule in the next layer so that they get a cleaner input, thus determining the pose of the objects more accurately. 

The right panel of Fig. \ref{capsule} shows a lower-level capsule (e.g. nose capsule) making decision to send its output to the parent capsules. This decision is made by adjusting the routing coefficients, $c_{ij}$, that will be multiplied by the prediction vectors before sending it to high level capsules. CapsNets compute the parent capsules and routing coefficients as follows:
\begin{gather}
    \mathbf{v}_j = \frac{||\mathbf{s}_j||^2}{1+||\mathbf{s}_j||^2}\frac{\mathbf{s}_j}{||\mathbf{s}_j||}, \quad
    \mathbf{s}_j=\sum_i c_{ij}\mathbf{\hat{u}}_{j|i}, \\ 
    c_{ij} = \frac{\text{exp}(b_{ij})}{\sum_k \text{exp}(b_{ik})}, \quad b_{ij}\leftarrow b_{ij} + \hat{\mathbf{u}}_{j|i}.\mathbf{v}_j.
\end{gather}
The output of each parent capsule $\mathbf{v}_j$ is computed as the weighted sum of all predictions from child capsules, then passed through a \textit{squash} non-linearity. Squashing makes sure that the output vector has length no more than 1 (so that its length can be interpreted as the probability that a given feature being detected by the capsule) without changing its direction. Parent capsules receive predictions from all children. These vectors are represented by points in Fig.~\ref{capsule}. The dynamic routing mechanism will increase the routing coefficient to  $j$-parent capsule by a factor of $\mathbf{\hat{u}}_{j|i}.\mathbf{v}_j$. Thus a child capsule will send more information to the parent capsule whose output $\mathbf{v}_j$ is more similar to its prediction $\mathbf{\hat{u}}_{j|i}$.

\noindent
\textbf{Network Architecture}:
The network contains two main parts: \textit{encoder} and \textit{decoder}, depicted in the first two figures of \cite{sabour2017dynamic}. The encoder contains three layers: two convolution and one fully-connected. The first layer is a standard convolution layer with 256 filters of size $9 \times 9$ and stride 1, followed by ReLU activation. The next layer is a convolutional capsule layer called the \textit{PrimaryCaps} layer. Capsules are arranged in 32 channels where each primary capsule applies 8 convolutional filters of size $9 \times 9$ and stride 2 to the whole input volume. In this setting, all PrimaryCaps in each 32 channels share their weights with each other and each capsule outputs an 8-dimensional vector of activations. The last layer is called \textit{DigitCaps} layer which has one 16D capsule per class. Routing takes place in between these capsules and all PrimaryCaps, encoding the input into 16-dimensional activation vector of instantiation parameters. The lengths of these prediction vectors are used to determine the predicted class. 

The decoder tries to reconstruct the input from the final capsules, which  will force the network to preserve as much information from the input as possible across the whole network. This effectively works as a regularizer that reduces the risk of over-fitting and helps generalize to new samples. In decoder, the 16D outputs of the final capsules are all masked out (set to zero) except for the ones corresponding to the target (while training) or predicted (while testing) class. They proposed using a three layer feed forward neural network with 512, 1024, and 784 units to reconstruct the input image.

\section{Fast Capsule Network}
 \textbf{Consistent Dynamic Routing}: Dynamic routing has been showed to improve the classification accuracy of CapsNet \cite{sabour2017dynamic}. However this operation is computationally expensive and does not scale well to high dimensional data,  such as 3D CT scans with a large number of pixels. We propose a consistent dynamic routing mechanism to address this problem. Specifically, we enforce all capsules in the PrimaryCaps layer corresponding to the same pixels to have the same routing coefficients:
 \begin{equation}
     c_{ij} = c_{kj}, \;\; \forall i,k \in \mathcal{S}= \{i, k \; | \; loc(i) = loc(k)\}
     \label{eq:consistent_routing}
 \end{equation}
 where $loc()$ is the function converting a capsule index to its pixel location. This strategy will dramatically reduce the number of routing coefficients. For example, the original CapsNet has 32 capsules for each pixel location in PrimaryCaps layer. The proposed method will therefore reduce the number of routing coefficients by a factor of 32. The iterative routing procedure is a bottleneck of CapsNets. Through reducing the number of routing operations, we expect to dramatically increase the network's efficiency. There are many possible ways to enforce the constraint in Eq.~(\ref{eq:consistent_routing}). In this paper, we use a simple strategy that allows only one capsule at each pixel location. To compensate for the reduction of number of capsules, we increase the dimension of each capsule to 256D instead of 8D as in the original CapsNet. Our results show that the proposed architecture achieves the similar classification accuracy, and is $3\times$ faster than CapsNet.
 
\begin{figure}
    \centering
    \includegraphics[width=\textwidth]{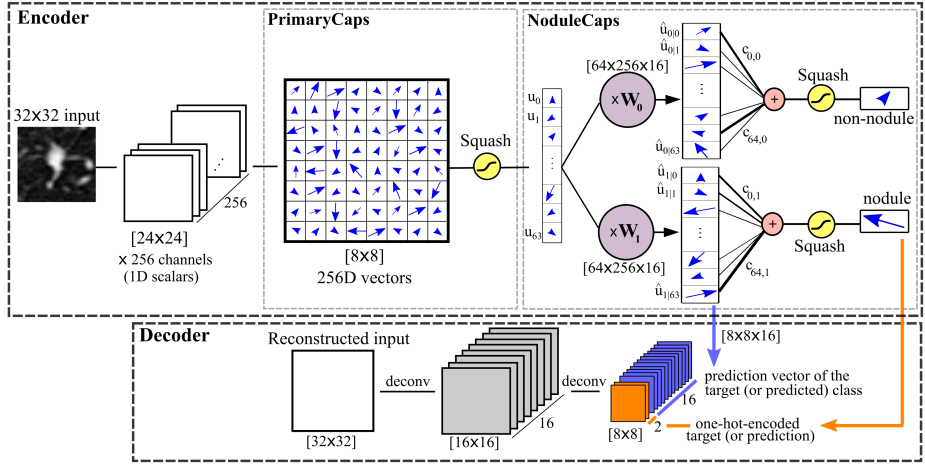}
    \caption{Visual representation of Fast Capsule Network}
    \label{fig:net}
\end{figure}

\noindent
\textbf{Convolutional Decoder}: CapsNet uses an additional reconstruction loss as a regularization to prevent over-fitting during learning the network's parameters \cite{sabour2017dynamic}. This encourages the digit capsules to encode as much information from the input as possible. The reconstruction is done by feeding 16D output of the final capsules to a three-layer feed-forward neural network. Unfortunately, our experiments show that lung nodule data cannot be reconstructed well from the final capsules. This could be due to the high visual variability within the data. To this end, we propose a different reconstruction mechanism to serve as the training regularization. Instead of using the masked outputs of the NoduleCaps, we used the PrimaryCaps prediction vectors ($\mathbf{\hat{u}}_{j|i}$) only for the target (or predicted) class. For the example in Fig.~\ref{fig:net}, it is $8 \times 8 \times 16$ tensor. We also concatenated the one-hot-encoded target (predicted) vector replicated over the space to form a final tensor of $8 \times 8 \times 18$ as the decoder's input. This tensor is then fed to two consecutive fractionally-strided convolution layers (also known as deconvolution \cite{zeiler2010deconvolutional} or transposed convolution layers) with 16 and 1 filters of size $4 \times 4$ and stride 2. This structure have much fewer parameters (about 5K) and significantly outperforms the feed-forward network in the reconstruction task. Fig~\ref{fig:sample} shows examples of reconstructed images using CapsNet and the proposed method. Similar to \cite{sabour2017dynamic}, we used margin loss for classification and L2 for reconstruction.

\section{Experiments}
\textbf{Dataset}: The study included 226 unique Computed Tomography (CT) Chest scans (with or without contrast) captured by General Electric and Siemens scanners. The data was preprocessed by an automated segmentation software in order to identify structures to the organ level. From within the segmented lung tissue, a set of potential nodule point is generated based on the size and shape of regions within the lung which exceeds the air Hounsfield Unit (HU) threshold. Additional filters, based on symmetry and other common morphological characteristics, are applied to decrease the false positive rate while maintaining very high sensitivity.

\begin{figure}[!t]
\centering
\includegraphics[width=\textwidth]{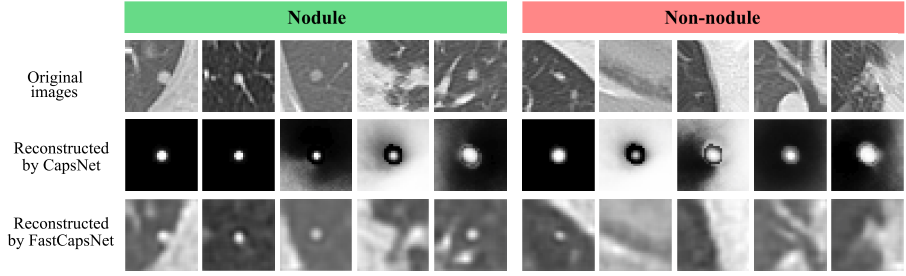}
\caption{Sample images of nodules (left) and non-nodules (right) and their reconstructions using the original CapsNet (middle row) and the FastCapsNet (last row)}
\label{fig:sample}
\vspace{-8mm}
\end{figure}

Bounding boxes with at least 8 voxels padding surrounding the candidate nodules are cropped and resized to $32\times 32\times 32$ pixels. Each generated candidate is reviewed and annotated by at least one \emph{board certified radiologist}. From all the generated images (about 7400 images), around 56\% were labeled as nodules and the rest non-nodules. The first row of Fig. \ref{fig:sample} shows examples of extracted candidates and the corresponding labels provided by radiologists. These images illustrate the highly challenging task of distinguishing nodules from non-nodule lesions. One reason is that the pulmonary nodules come with large variations in shapes, sizes, types, etc. The second reason which hinders the identification process is the non-nodule candidates mimicking the morphological appearance of the real pulmonary nodules. For all these reasons, the detection and classification of lung nodules is a challenging task, even for experienced radiologists.

\noindent
\textbf{2D CapsNet}: Fig. \ref{fig:net} provides an illustration of the proposed architecture, which consists of 3 different layers. To prepare the required 2D images, we used the middle slices of the 3D volumes along the x-axis (as x-axis contains more information according to radiologist's feedback). The first layer is a 3D convolution layer with 256 filters of size 9 and stride 1. Given our $32\times 32$ pixel inputs, this layer outputs  $24\times 24\times 256$ tensor which are the basic features extracted from input image. The next layer is the \textit{PrimaryCaps} layer which applies 256 convolutional filters on the input to obtain $8\times 8\times 256$ tensor. We consider this tensor to be $8\times 8 = 64$ capsules, each with 256D. This results in only one capsule at each pixel, therefore effectively enforces consistent dynamic routing. These vectors will then be passed through a squashing non-linearity to get the final output of the lower-level capsules (\textit{i.e.} $\mathbf{u}_i$). The proposed architecture is a departure from from the original CapsNet, which divides the tensor into $8\times 8\times 32 = 2048$ capsules, each of 8D, resulting in 32 capsules for each pixel locations. While the total number of parameters of the two networks are the same, our network drastically reduces the number of PrimaryCaps from 2048 to 64, thereby decreases the number of voting coefficients by 32 times. 

In the final layer, called \textit{NoduleCaps}, 256D output from PrimaryCaps layer are multiplied with its own $256\times16$ transformation matrix which maps the output to the 16D space (results in $\mathbf{\hat{u}}_{j|i}$). Finally, the routing parameter tunes the coupling coefficients and decides about the amount of contribution they make to the NoduleCaps. During training, the coefficients are initialized uniformly, meaning that the output from the PrimaryCaps layer is sent equally to all the capsules in the NoduleCaps layer. Then the routing between the two layers is learned iteratively.

\begin{table}[b]
\vspace{-2mm}
\centering
\caption{Comparison of the performance, number of trainable parameters (M is for millions), and required training time per epoch of the Deep Networks}
\vspace{-2mm}
\label{table:result}
\begin{tabular}{lccccc}
\hline
\multicolumn{1}{c}{} & \;\;\textbf{precision}\;\; & \;\;\textbf{recall}\;\; & \;\;\textbf{error rate}\;\; & \;\;\textbf{\#params}\;\; & \;\textbf{sec/epoch}\; \\ \hline
Automated Software \; & 73.65      & 82.13  & 25.34     & -     & -     \\ \hline
2D AlexNet                 & 88.09    & 85.52  & 13.09     & 7.6M  & 6.9   \\
2D ResNet-50               & 89.14    & 85.52  & 12.51     & 23.5M & 16.3  \\ 
2D CapsNet              & 89.44    & 85.98  & 12.16     & 7.4M  & 18.8  \\
2D FastCapsNet          & \textbf{89.71}    & \textbf{87.41}  & \textbf{11.45}     & 5.8M  & 6.2   \\ \hline
3D AlexNet              & 91.05 & 87.92 & 10.42 & 7.8M & 175.1 \\
3D ResNet-50              & 90.51 & \textbf{90.42} & 9.58 & 48.2M & 341.5     \\
3D FastCapsNet          & \textbf{91.84} & 89.11 & \textbf{9.52} & 52.2M & 320.0   \\ \hline
\end{tabular}
\vspace{-5mm}
\end{table}

\noindent
\textbf{3D Capsule Network}: The 3D version of the proposed architecture is structurally similar to the explained 2D version. In the first two layers, 2D convolutions are replaced by 3D convolution operators capturing the cross-correlation of the voxels. The number and size of the filters and the strides are the same as before. This results in an $8 \times 8 \times 8$ volume of primary capsules, each sees the output of all $256\times9\times9\times9$ units of the previous convolution layer. Therefore, our proposed faster architecture gives 512 PrimaryCaps, compared with 23328 of the original network. This will limit the number of required routing coefficients, help the dynamic routing perform faster and perform significantly better. NodulesCaps are the same as in the 2D network, fully-connected to PrimaryCaps with the dynamic routing happening in the middle. Similar to the proposed 2D FastCapsNet, the decoder takes the prediction vectors of all \textit{PrimaryCaps} (only the ones routed to the correct class), concatenates it with the one-hot-encoded label of the target to form a tensor of shape $8\times8\times8\times18$. It then reconstruct the input from this tensor by applying two consecutive 3D fractionally-strided convolution layers with 16 and 1 filters, respectively. Obviously, a feed-forward architecture wouldn't scale well for reconstructing the 3D images (it requires $32\times32\times32$ output units which drastically increases the number of parameters).

\begin{figure}[t]
    \centering
    \includegraphics[width=\textwidth]{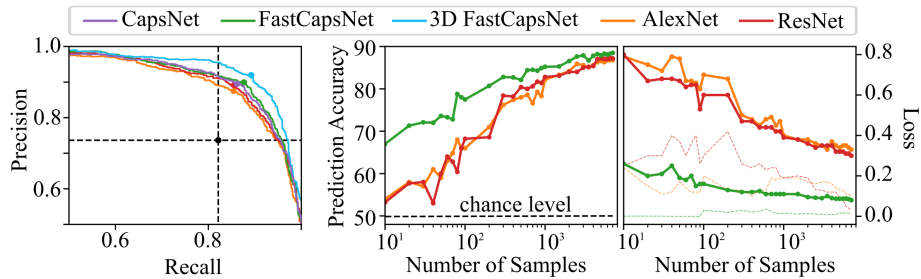}
    \caption{\textbf{Left:} test precision-recall curves, \textbf{Center:} test prediction accuracy of models trained on different number of training samples, \textbf{Right:} train (dashed) and test (solid) loss of models trained on different number of training samples.}
    \label{fig:result}
    \vspace{-5mm}
\end{figure}

\noindent
\textbf{Results}:
2D version of the proposed faster CapsNet was trained and compared with two well-known deep network architectures, AlexNet \cite{krizhevsky2012imagenet} and ResNet-50 \cite{he2016deep}, as well as the original Capsule network. Trying both 2D and 3D architectures helps testing the network's performance in both settings, as well as enabling us to quantify the contribution of the third dimension. All analyses are done using a desktop machine equipped with 128 GB of memory, 4 NVidia GTX 1080 with 8 GB of video memory each, and Intel Xeon CPU 2.40GHz with 16 cores.

The test prediction results are presented in Table \ref{table:result}. Capsule networks achieve relatively lower error rates compared to AlexNet and ResNet. FastCapsNet has fewer parameters than other models, thus is more than 3 times faster than the regular CaspNet in both training and test runs. For 3D images, the 3D implementation of the original CapsNet failed to converge. The prediction accuracy remains close at the chance level with the validation loss increasing gradually. However, our proposed 3D FastCapsNet converges relatively fast (reaching less than 20\% test loss after only 3 training epochs). 

The precision-recall curve of the models are presented in the left panel of Fig. \ref{fig:result} confirming the superior performance of the proposed FastCapsNets compared to the other models. The 3D FastCapsNet performs significantly better than the rest, beating the 2D version by almost 2\% accuracy indicating the contribution of the third dimension.
We also trained the 2D models using different number of training samples. The prediction accuracy and loss values of different models are presented in the middle and right panel of Fig. \ref{fig:result}, respectively. Capsule network performs comparatively more robust. For very few number of samples, AlexNet and ResNet-50 go almost to chance level while capsule network stays about 70\%.
\vspace{-6mm}
\section{Conclusions}
\vspace{-2mm}
Our work shows that CapsNet is a promising alternative to CNN. Experimental results demonstrate that CapsNets compare favorably to CNNs when the training size is large, but significantly outperform CNNs on small size datasets. We showed that, by modifying the routing mechanism, we can speed up CapsNets 3 times while maintaining the same accuracy. In the future, we will explore unsupervised learning with CapsNets.

{\small
\bibliographystyle{ieeetr}
\bibliography{references.bib}
}
\end{document}